# Multiple instance learning for sequence data with across bag dependencies

Manel Zoghlami [1, 2, *], Sabeur Aridhi [3], Mondher Maddouri [4], Engelbert Mephu Nguifo [1]

[1] Clermont Auvergne University, CNRS, LIMOS, France
[2] University of Tunis El Manar, Faculty of sciences of Tunis, LIPAH, Tunisia
[3] University of Lorraine, CNRS, Inria, LORIA, France
[4] University of Jeddah, College Of Business, KSA

## Abstract
In Multiple Instance Learning (MIL) problem for sequence data, the learning data consist of a set of bags where each bag contains a set of instances/sequences. In some real world applications such as bioinformatics, comparing a random couple of sequences makes no sense. In fact, each instance may have structural and/or functional relations with instances of other bags. Thus, the classification task should take into account this across bag relation. In this work, we present two novel MIL approaches for sequence data classification named ABClass and ABSim. In ABClass, motifs are extracted from related instances and used in order to encode sequences. A discriminative classifier is then applied to compute a partial classification result for each set of related sequences. In ABSim, we use a similarity measure to discriminate the related instances and then build a scores matrix. For both approaches, an aggregation method is applied in order to generate the final classification result. We applied both approaches to solve the problem of bacterial Ionizing Radiation Resistance (IRR) prediction. The experimental results of the presented approaches are satisfactory.

**Availability:** Additional information about our datasets and our experiments can be found in the following link: *http://homepages.loria.fr/SAridhi/software/MIL/*

**Keywords:** sequence data classification, multiple instance learning, bacterial ionizing radiation resistance.

## 1. Introduction

In a traditional setting of machine learning, each instance is associated with a label. However, in Multiple Instance Learning (MIL), we have a set of bags where each bag contains multiple instances and it is the bag that carries a label. The labels of the instances that conform the bag could be unknown.

This work was originally proposed to solve the problem of bacterial Ionizing Radiation Resistance (IRR) prediction (Zoghlami *et al*. 2018; Aridhi *et al*. 2016). Each bacterium is represented by a set of protein sequences. This problem could be formalized as an MIL problem: bacteria represent the bags and protein sequences represent the instances. Each protein sequence may differ from a bacterium to another, e.g., each bag contains the protein named *Endonuclease III,* but it is expressed differently from one bag to another: these are called orthologous proteins (Fang *et al*. 2010). In order to learn the label of an unknown bacterium, comparing a random couple of sequences makes no sense, it is rather better to compare the protein sequences that have a functional relationship/dependency: the orthologous proteins. Hence, this work deals with the MIL problem that has the following three criteria:

1. **The instances inside the bags are sequences:** to deal with sequences, we have to deal with data representation. A widely used technique to represent MIL sequence data is to apply a preprocessing step which extracts features/motifs to represent the sequences (Sutskever *et al*. 2014; Saidi *et al*. 2010; Lesh *et al*. 1999). Other works keep data in their original format and use sequence comparison techniques such as defining a distance function to measure the similarity between pairs of sequences (Aridhi *et al*. 2016; Xing *et al*. 2010; Saigo *et al*. 2004).

* Corresponding author – E-mail: manel.zoghlami@etu.uca.fr

**2. All the instances inside a bag contribute to define the bag's label:** the standard MIL assumption states that every positive bag contains at least one positive instance while in every negative bag all of the instances are negative. This is not guaranteed to hold in some domains so alternative assumptions are considered (Foulds and Frank 2010). Particulary, the standard assumption is not suitable for the above bacterial IRR prediction problem since one positive instance is not sufficient to classify a bag as positive. All the protein sequences have to contribute to the final decision. We opt for the collective assumption that considers that all of the instances contribute to the bag's label (Amores 2013; Foulds and Frank 2010).

**3. The instances may have dependencies across the bags:** one major assumption of most existing MIL methods is that each bag contains a set of instances that are independently distributed. Nevertheless, in many applications, the dependencies between instances naturally exist and if incorporated in the classification process, they can potentially improve the prediction performance significantly (Zhang *et al*. 2011). In the case of the IRR prediction problem, the bags contain orthologous protein sequences. Considering this issue, the problem we want to solve in this work is an MIL problem in sequence data that have dependencies between instances of different bags.

In this work, we present two approaches named ABClass and ABSim. ABClass performs a preprocessing step that extracts motifs from each set of related sequences in order to use them as attributes to construct a vector representation for each set of sequences. In order to compute partial prediction results, a discriminative classifier is applied to each sequence of the unknown bag and its correspondent related sequences in the learning dataset. Finally, an aggregation method is applied to generate the final result. In ABSim, we use a similarity measure between each sequence of the unknown bag and the corresponding sequences in the learning bags in order to create a similarity scores matrix. An aggregation method is then applied and the unknown bag is labeled according to the bag that presents more similar sequences.

The remainder of this paper is organized as follows. Section 2 provides a formalization of the problem of MIL in sequence data followed by a simple use case that serves as a running example throughout this paper. In Section 3, we present an overview of some related works in MIL. Section 4 gives a description of a naive approach followed by a detailed description of ABClass and ABSim approaches. In Section 5, we applied the proposed approaches to solve the problem of IRR prediction in bacteria. We describe our experimental environment and we discuss the obtained results. Concluding points make the body of Section 6.

# 2. Problem formulation

We denote $\Sigma$ an alphabet defined as a finite set of characters or symbols. A sequence is defined as an ordered list of symbols from $\Sigma$. Let $DB$ be a learning database that contains a set of n labeled bags $DB = \{(B_i, Y_i), i = 1, \ldots, n\}$ where $Y_i=\{-1, 1\}$ is the label of the bag $B_i$ . Instances in $B_i$ are sequences and are denoted by $B_{ij}$. Formally $B_i = \{B_{ij}, j = 1, \ldots, m_{Bi}\}$, where $m_{Bi}$ is the total number of instances in the bag $B_i$. We do not have necessarily the exact number of instances in each bag. The problem investigated in this work is to learn a multiple instance classifier from $DB$. Given a query bag $Q = \{Q_k, k = 1, \ldots, q\}$ where q is the total number of instances in $Q$, the classifier should use sequential data in this bag and in each bag of $DB$ in order to predict the label of $Q$.

We note that there is a relation $R$ between instances of different bags denoted *the across bag relation*. To represent this relation, we opt for an index representation. In fact, a preprocessing step assigns an index number to the instances inside each bag according to the following notation: each instance $B_{ij}$ of a bag $B_i$ is related by $R$ to the instance $B_{hj}$ of another bag $B_h$ in $DB$.

$$R: DB \rightarrow DB$$
$$R(B_{ij}) = B_{hj}$$

where $i$ and $h \in \{1, ..., n\}$, $i \neq h$ and $j \in \{1, ..., m\}$. We note that this notation does not mean that instances are ordered. An instance may not have any corresponding related instance in some bags, i.e., a sequence is related to zero or one sequence per bag. The relation $R$ could be generalised to deal with problems where each instance has more than one target related instance in each bag. The index notation as described previously will not be suitable in this case.

In order to illustrate our proposed approaches, we rely on the following running example. Let $\Sigma = \{A, B, ..., Z\}$ be an alphabet. Let $DB = \{(B_1, +1), (B_2, +1), (B_3, -1), (B_4, -1), (B_5, -1)\}$ a learning database that contains five bags ($B_1$ and $B_2$ are positive bags, $B_3$, $B_4$ and $B_5$ are negative bags). Initially, the bags contain the following sequences:
$B_1$ = {ABMSCD, EFNOGH, RUVR}
$B_2$ = {CCGHDDEF, EABZQCD}
$B_3$ = {GHWMY, ACDXYZ}
$B_4$ = {ABIJYZ, KLSSO, EFYRTAB}
$B_5$ = {EFFVGH, KLSNAB}

We first use the across bag relation $R$ to represent the related instances using the index notation as described previously.

$$B_1 = \begin{cases} B_{11} = \mathbf{AB}MS\mathbf{CD} \\ B_{12} = \mathbf{EF}N\,O\mathbf{GH} \\ B_{13} = RUVR \end{cases} \quad B_2 = \begin{cases} B_{21} = E\mathbf{AB}ZQ\mathbf{CD} \\ B_{22} = CC\mathbf{GH}DD\mathbf{EF} \end{cases} \quad B_3 = \begin{cases} B_{31} = A\mathbf{CD}X\mathbf{YZ} \\ B_{32} = \mathbf{GH}WMY \end{cases}$$

$$B_4 = \begin{cases} B_{41} = \mathbf{AB}IJ\mathbf{YZ} \\ B_{42} = \mathbf{EF}Y\,RT\mathbf{AB} \\ B_{43} = \mathbf{KL}SSO \end{cases} \quad B_5 = \begin{cases} B_{52} = \mathbf{EF}FV\mathbf{GH} \\ B_{53} = \mathbf{KL}SN\mathbf{AB} \end{cases}$$

We need to predict the class label of an unknown bag $Q = \{Q_1, Q_2, Q_3\}$ where:

$$Q = \begin{cases} Q_1 = \mathbf{AB}WX\mathbf{CD}Y \\ Q_2 = \mathbf{EF}XY\mathbf{GH}NV \\ Q_3 = \mathbf{KL}OF \end{cases}$$

## 3. Multiple Instance Learning (MIL)

In a traditional setting of machine learning, the training set is composed of instances associated with labels. In an MIL setting, we have a set of bags where each bag contains multiple instances and it is the bag that carries a label. Several MIL approaches have been proposed. Some algorithms deal with the MIL problem directly in either instance level such as mi-SVM (Andrews *et al*. 2003) and MILKDE (Faria *et al*. 2017) or in bag level such as MI-SVM (Andrews *et al*. 2003) and MIGraph (Zhou *et al*. 2009). Other algorithms try to shift the MIL problems into instance space via embedding such as MILDE (Amores 2015), Submil (Yuan *et al*. 2016) and miVLAD (Wei *et al*. 2017). In what follows, we will present mainly some MIL algorithms used with the naive approach in our experiments and some works that apply MIL algorithms on multiple instance sequence datasets. A review of MIL approaches and a comparative study could be found in (Amores 2013), (Alpaydin *et al*. 2015) and (Herrera *et al*. 2016). The Diverse Density (DD) (Maron and Pérez 1998) algorithm is proposed as a general framework for solving MIL problems. It attempts to find the concept points in the feature space that are close to at least one instance from every positive bag and far from instances in negative bags. The optimum concept point can be found by maximizing the diversity density score, which is a measure of how many different positive bags have instances near the point and how far the negative instances are away from it. In (Wang 2000), the authors present two extensions of the kNN algorithm called Bayesian-kNN and Citation-kNN. In order to classify an unknown bag, the Bayesian method computes the posterior probabilities of its label based on the labels of its neighbors. Citation-kNN algorithm classifies a bag based on the labels of its references (the nearest neighbors bags) and citers (the bags that count the processed bag as their neighbor). MI-SVM and mi-SVM (Andrews *et al*. 2003) are two extensions of the support vector machines where margin maximization is redefined in order to consider the MIL settings. mi-SVM works at instance level and uses a maximum instance margin formulation which tries to recover the instance labels of the positive bags. On the other hand, MI-SVM generalizes the notion of a margin to bags. The margin of a positive bag is defined by the margin of the most positive instance, while the margin of a negative bag is defined by the least negative instance.

Using the attribute-value format in order to encode the input data is widely used when applying MIL algorithms on sequence data (Zhang *et al*. 2011; Tao *et al*. 2004; Andrews *et al*. 2003). In (Andrews *et al*. 2003), the empirical evaluation uses the TREC9 dataset for document categorisation. Terms are used to present the text. This leads to an extremely sparse and high dimensional attribute-value representation of the data. In (Tao *et al*. 2004; Zhang *et al*. 2011), a dataset of protein sequences is used in the empirical evaluation. The goal is to identify *Trx-fold proteins*. Each primary protein sequence is considered as a bag, and some of its subsequences are considered as instances. These subsequences are aligned and mapped to an 8-dimensional feature space: 7 numeric properties (Kim *et al*. 2000) and an $8^{th}$ feature that represents the residue's position. This results to an attribute-value format description of the dataset.

# 4. MIL approaches for sequence data with across bag dependencies

In this section, we first describe the naive approach for MIL in sequence data. Then, we present our proposed approaches.

## 4.1. Naive approach for MIL in sequence data with across bag dependencies

The simplest way to solve the problem of MIL for sequence data is to use standard MIL classifiers. The naive approach contains two steps (see Figure 1). We first make a preprocessing step that transforms the set of sequences to an attribute-value matrix where each row corresponds to a bag of sequences and attributes conform the columns. The second step consists in applying an existing MIL classifier. In the case of sequence data, the most used technique to transform data to an attribute-value format is to extract motifs that serve as attributes/features. We note that finding a uniform description of all instances using a set of motifs is not always an easy task. Since our naive approach takes into account the across bag relations between instances, the preprocessing step extracts motifs from each set of related instances. The union of these extracted motifs is then used as features to construct an attribute-value matrix where each row corresponds to a bag. The presence or the absence of an attribute in a sequence is respectively denoted by 1 or 0. Using this approach, we obtain an attribute-value matrix that contains a large number of motifs. It is worthwhile to mention that only a subset of the used attributes is representative for each processed sequence. Therefore, we may have a big sparse matrix when trying to present the whole sequence data using an attribute value format.

We apply the naive approach to our running example. We suppose that attributes are subsequences (minimum length = 2) that occur at least in 2 instances. Let $AttributeList_1$ = {AB, CD, YZ} be the list of features extracted from the instances $\{B_{i1},\ i=1,\ \ldots,\ 4\}$. $AttributeList_2$ = {EF, GH} is the list of features extracted from the instances $\{B_{i2};\ i = 1,\ \ldots,\ 5\}$ and $AttributeList_3$ = {KL} is the list of features extracted from the instances $\{B_{i3},\ i \in \{1, 4, 5\}\}$. The union of the extracted motifs produces the list *AttributeList* = {AB, CD, YZ, EF, GH, KL}. In order to encode the learning sequence data, we generate the following attribute-value matrix denoted ***M***:

$$
\begin{array}{c}
\qquad\quad 1^{st}\ \text{instance} \qquad\qquad\qquad 2^{nd}\ \text{instance} \qquad\qquad\qquad 3^{rd}\ \text{instance} \\
\mathbf{M} = \left(\begin{array}{cccccc|cccccc|cccccc}
1&1&0&0&0&0&0&0&0&1&1&0&0&0&0&0&0&0\\
1&1&0&0&0&0&0&0&0&1&1&0&-&-&-&-&-&-\\
0&1&1&0&0&0&0&0&0&0&1&0&-&-&-&-&-&-\\
1&0&1&0&0&0&1&0&0&1&0&0&0&0&0&0&0&1\\
-&-&-&-&-&-&0&0&0&1&1&0&1&0&0&0&0&1
\end{array}\right)
\begin{array}{l} B_1\\ B_2\\ B_3\\ B_4\\ B_5 \end{array}
\end{array}
$$

The sparsity percentage of *M* is 77.2%. If we have a big learning database, *M* could result to a huge and sparse matrix since only a subset of the used subsequences is representative for each processed sequence.

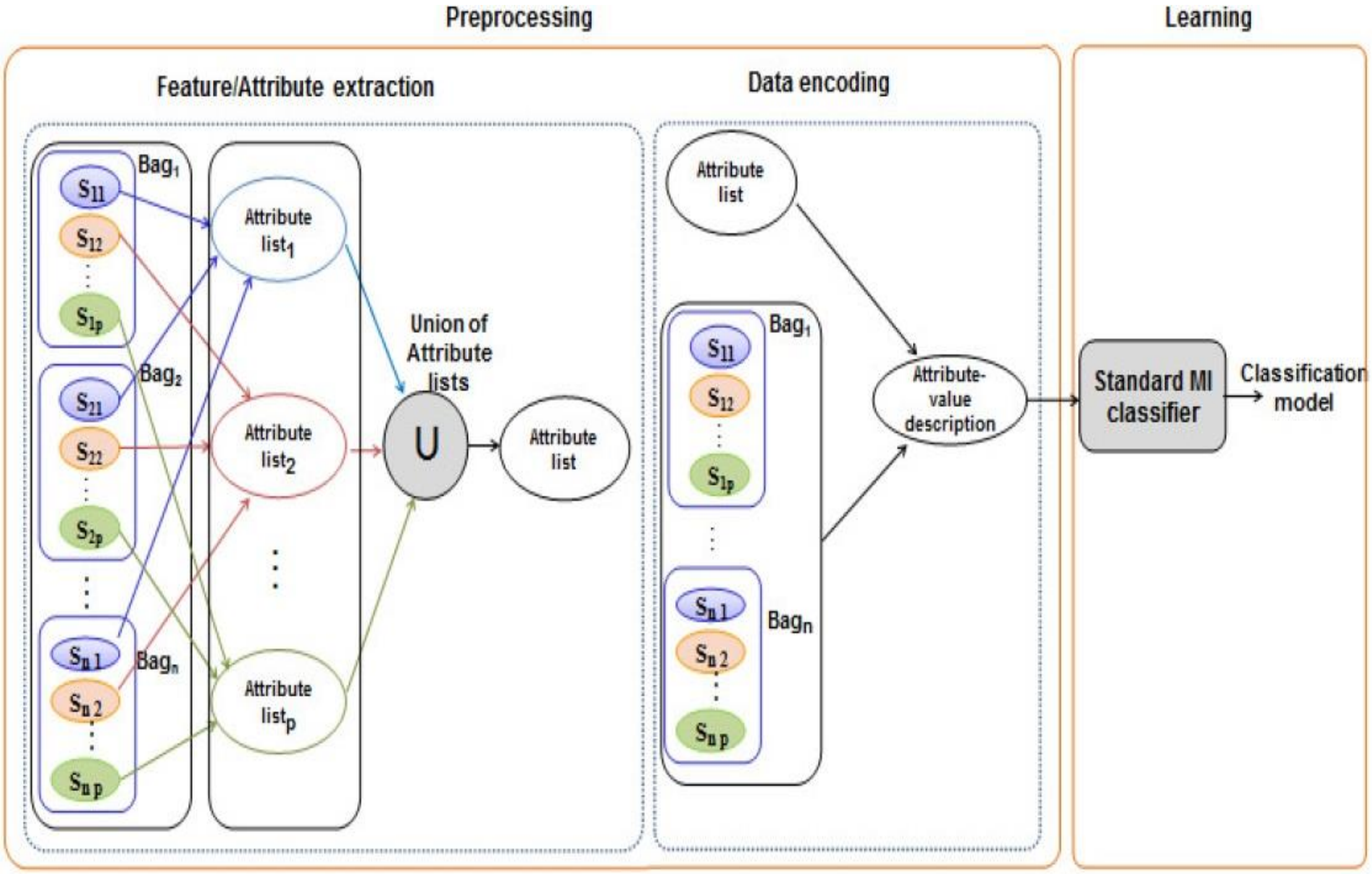

***Figure 1 System overview of the naive approach for MIL in sequence data***

## 4.2. ABClass: Across Bag sequences Classification approach

ABClass takes advantage of the across bag relationship between sequences in order to reduce the number of attributes that are not representative for each processed sequence during the encoding step. Figure 2 represents the system overview of ABClass. Each set of related instances is presented by its own motifs vector. This relationship is also used during the learning step when generating partial models. Every vector of motifs will be used to produce a partial prediction result. These results will be then aggregated to compute the final result. Based on the formalization, the algorithm discriminates bags by applying a classification model to each instance of the query bag.

**Algorithm 1** *ABClass*

**Input:** Learning database $DB = \{(B_i, Y_i)\ , i= 1, \ldots, n\}$, Query bag $Q = \{Q_k, k= 1, \ldots, q\}$
**Output:** Prediction result $P$
1: **for all** $Q_k \in Q$ **do**
2: $AcrossBagSeqList_k \leftarrow AcrossBagSeq\ (k, DB)$
3: $MotifList_k \leftarrow MotifExtractor\ (AcrossBagsList_k)$
4: $M_k \leftarrow EncodeData\ (MotifList_k, AcrossBagsList_k)$
5: $Model_k \leftarrow GenerateModel\ (M_k)$
6: $QV_k \leftarrow EncodeData\ (MotifList_k, Q_k)$
7: $PV_k \leftarrow ApplyModel\ (QV_k, Model_k)$
8: **end for**
9: $P \leftarrow$ *Aggregate* (PV)
10: **return** $P$

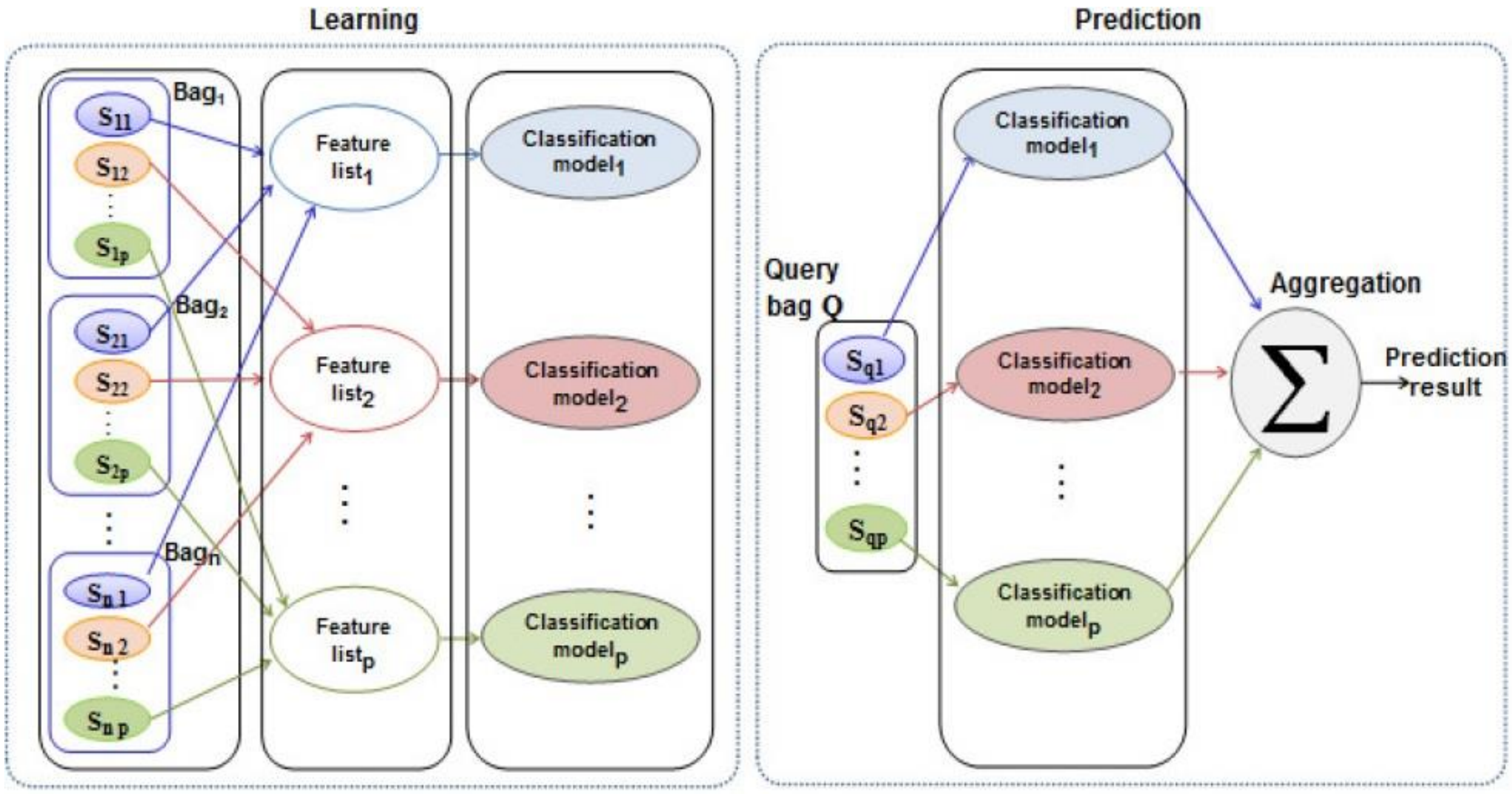

***Figure 2 System overview of the ABClass approach***

ABClass is described in Algorithm 1. The *acrossBagSeq* function groups the related instances among bags into a list. During the execution of the algorithm, we will use the following variables:

- A matrix *M* to store the encoded data of the learning database.
- A vector *QV* to store the encoded data of the query bag.
- A vector *PV* to store the partial prediction results.

Informally, the main steps of the ABClass algorithm are:

1. For each instance sequence $Q_k$ in the query bag *Q*, the related instances among bags of the learning database are grouped into a list (lines 1 and 2).
2. The algorithm extracts motifs from the list of grouped instances. These motifs are used to encode instances in order to create a discriminative model (lines 3 to 5).
3. ABClass uses the extracted motifs to represent the instance $Q_k$ of the unknown bag into a vector $QV_k$, then it compares it with the corresponding model. The comparison result is stored in the $k^{th}$ element of a vector *PV* (lines 6 and 7).

4. An aggregation method is applied to *PV* in order to compute the final prediction result *P* (line 9), which consists in a positive or a negative class label.

We apply the ABClass approach to our running example. Since the query bag contains 3 instances $Q_1, Q_2$ and $Q_3$, we need 3 iterations followed by an aggregation step.

**Iteration 1:** ABClass groups the set of bags that are related and extract the corresponding motifs.

$$AcrossBagsList_1 = \{B_{11}, B_{21}, B_{31}, B_{41}\}$$
$$MotifList_1 = \{AB, CD, YZ\}$$

Then, it generates the attribute-value matrix $M_1$ describing the sequences related to $Q_1$.

$$\begin{array}{cc} & \begin{array}{ccc} AB & CD & YZ \end{array} \\ \mathbf{M_1=} & \begin{pmatrix} \mathbf{1} & \mathbf{1} & \mathbf{0} \\ \mathbf{1} & \mathbf{1} & \mathbf{0} \\ \mathbf{0} & \mathbf{1} & \mathbf{1} \\ \mathbf{1} & \mathbf{0} & \mathbf{1} \end{pmatrix} \begin{array}{c} B_{11} \\ B_{21} \\ B_{31} \\ B_{41} \end{array} \end{array}$$

The sparsity percentage of the produced matrix $M_1$ is reduced to 33% because there is no need to use the motifs extracted from instances $\{B_{i2}, i = 1, .., 5\}$ and $\{B_{i3}, i \in \{1, 4, 5\}\}$ to describe the instances $\{B_{i1}, i = 1, .., 4\}$. A model is then created using the encoded data and a vector $QV_1$ is generated to describe $Q_1$.

$$\mathbf{QV_1=} \begin{pmatrix} \mathbf{1} \\ \mathbf{1} \\ \mathbf{0} \end{pmatrix}$$

By applying the model to the vector $QV_1$, we obtain the first partial prediction result and we store it into the vector *PV*.

$$PV_1 \leftarrow ApplyModel\ (QV_1, Model_1)$$

**Iteration 2:** The second iteration concerns the second instance $Q_2$ of the query bag. We do the same instructions described in the first iteration.

$$AcrossBagsList_2 = \{B_{12}, B_{22}, B_{32}, B_{42}, B_{52}\}$$
$$Motif\ List_2 = \{EF, GH\}$$

$$\begin{array}{cc} & \begin{array}{cc} EF & GH \end{array} \\ \mathbf{M_2=} & \begin{pmatrix} \mathbf{1} & \mathbf{1} \\ \mathbf{1} & \mathbf{1} \\ \mathbf{0} & \mathbf{1} \\ \mathbf{1} & \mathbf{0} \\ \mathbf{1} & \mathbf{1} \end{pmatrix} \begin{array}{c} B_{12} \\ B_{22} \\ B_{32} \\ B_{42} \\ B_{52} \end{array} \end{array}$$

$$\mathbf{QV_2=} \begin{pmatrix} \mathbf{1} \\ \mathbf{1} \end{pmatrix}$$

$$PV_2 \leftarrow ApplyModel\ (QV_2, Model_2)$$

**Iteration 3:** Only $B_1$, $B_4$ and $B_5$ have related instances to $Q_3$.

$$AcrossBagsList_3 = \{B_{13}, B_{43}, B_{53}\}$$
$$Motif\ List_3 = \{KL\}$$

$$\begin{array}{cc} & KL \\ \mathbf{M_3=} & \begin{pmatrix} \mathbf{0} \\ \mathbf{1} \\ \mathbf{1} \end{pmatrix} \begin{array}{c} B_{13} \\ B_{43} \\ B_{53} \end{array} \end{array}$$

$$\mathbf{QV_3=(1)}$$

$$PV_3 \leftarrow ApplyModel\ (QV_3, Model_3)$$

The aggregation step is finally used to generate the final prediction decision using the partial prediction results. We opt for the majority vote.

## 4.3. ABSim: Across Bag sequences Similarity approach

According to the specificity of the processed data, a similarity measure can be defined and used to discriminate instances. We propose an algorithm that focuses on discriminating bags by measuring the similarity between each sequence in the query bag and its corresponding related sequences in the different bags of the learning database. ABSim was originally presented in (Aridhi et al. 2016) as an algorithm used for IRR prediction. When applied on genomic sequences, ABSim uses the alignment score as similarity measure to compare protein sequences.

**Algorithm 2** ABSim

**Input:** Learning database $DB = \{(B_i, Y_i), i= 1, \ldots, n\}$, Query bag $Q = \{Q_k, k = 1, \ldots, q\}$

**Output:** Prediction result $P$

1: **for all** $Q_k \in Q$ **do**

2:     **for all** $B_i \in DB$ **do**

3:         $M_{ik} \leftarrow$ *similarityMeasure* $(Q_k, B_{ik})$ /*$B_{ik}$ is the instance number $k$ in the bag $B_i$*/

4:     **end for**

5: **end for**

6: $P \leftarrow$ *Aggregate* $(M)$

7: **return** P

ABSim works as described in Algorithm 2. Let $M$ be a matrix used to store similarity measurement score vectors during the execution of the algorithm. Informally, the algorithm is described as follows:

1. For each instance sequence $Q_k$ in the query bag $Q$, it computes the corresponding similarity scores (line 1 to 5). The similarity scores of all instances of the query bag are grouped into a matrix $M$ (line 3). The element $M_{ik}$ corresponds to the similarity score between $Q_k$ of $Q$ and $B_{ik}$ of $B_i$.
2. An aggregation method is applied to $M$ in order to compute the final prediction result $P$ (line 6). According to the aggregation result, a class label is associated to the query Bag.

Two aggregation methods are defined: Sum of Maximum Scores (SMS) and Weighted Average of Maximum Scores (WAMS). Algorithms 3 and 4 illustrate the SMS and WAMS aggregation methods. For each sequence in the query bacterium, we scan the corresponding line of $M$, which contains the obtained scores against all the other bags of the training database. The SMS method selects the maximum score among the similarity scores against bags that belong to the positive class label (which we call $max_P$ ) and the maximum score among the similarity scores against bags that belong to the negative class label (which we call $max_N$). These scores are then compared. If $max_P$ is greater than $max_N$, it adds $max_P$ to the total score of the positive class label (which we denote $total_P (M)$). Otherwise, it adds $max_N$ to the total score of the negative class label (which we denote $total_N (M)$). When all selected sequences were processed, the SMS method compares total scores of positive class label and negative class label. If $total_P (M)$ is greater than $total_N (M)$, the prediction output is the positive class label. Otherwise, the prediction output is the negative class label.

Using the WAMS method, each sequence $Q_i$ has a given weight $w_i$. For each sequence in the query bag, we scan the corresponding line of $M$, which contains the obtained scores against all other bags of the training database. The WAMS method selects the maximum score among the similarity scores against bags that belong to positive class label (which we denote $max_P (M)$) and the maximum score among the similarity scores against bags that belong to the negative class label (which we denote $max_N (M)$). It then compares these scores. If the $max_P (M)$ is greater than $max_N (M)$, it adds $max_P (M)$ multiplied by the weight of the sequence to the total score of the positive class label and it increments the number of positive bags having a max score. Otherwise, it adds $max_N (M)$ multiplied by the weight of the sequence to the total score of the negative class label and it increments the number of negative bags having a max score. When all the selected sequences were processed, we compare the average of total scores of positive class labels (which we denote $avg_P (M)$) and the average of total scores of negative class labels (which we denote $avg_N(M)$). If $avg_P(M)$ is greater than $avg_N(M)$, the prediction output is the positive class label. Otherwise, the prediction output is the negative class label.

**Algorithm 3** SMS

**Input:** Similarity matrix $M = \{M_{ij}, i = 1, \ldots, n$ and $j = 1, \ldots, p\}$
**Output:** A prediction result $P$

$total_P \leftarrow 0$
$total_N \leftarrow 0$
**for** $i \in [1, n]$ **do**
    $max_P \leftarrow 0$
    $max_N \leftarrow 0$
    **for** $j \in [1, p]$ **do**
        **if** $Y_j = +1$ *and* $max_P \geq M_{ij}$ **then**
            $max_P \leftarrow M_{ij}$
        **else if** $Y_j = -1$ *and* $max_N \geq M_{ij}$ **then**
            $max_N \leftarrow M_{ij}$
        **end if**
    **end for**
    **if** $max_P \geq max_N$ **then**
        $total_P \leftarrow total_P + max_P$
    **else**
        $total_N \leftarrow total_N + max_N$
    **end if**
**end for**
**if** $total_P \geq total_N$ **then**
    $P \leftarrow +1$
**else**
    $P \leftarrow -1$
**end if**
**return** $P$

**Algorithm 4** WAMS

**Input:** Similarity matrix $M = \{M_{ij}, i = 1, \ldots, n$ and $j = 1, \ldots, p\}$, Weight vector $W = \{(W_i, i = 1, \ldots, p\}$
**Output:** A prediction result $P$

$total_P \leftarrow 0$, $total_N \leftarrow 0$, $nb_P \leftarrow 0$, $nb_N \leftarrow 0$
**for** $i \in [1, p]$ **do**
    $max_P \leftarrow 0$
    $max_N \leftarrow 0$
    **for** $j \in [1, n]$ **do**
        **if** $Y_j = +1$ *and* $max_P \geq M_{ij}$ **then**
            $max_P \leftarrow M_{ij}$
        **else if** $Y_j = -1$ *and* $max_N \geq M_{ij}$ **then**
            $max_N \leftarrow M_{ij}$
        **end if**
    **end for**
    **if** $max_P \geq max_N$ **then**
        $total_P \leftarrow total_P + (max_P \cdot w_i)$
        $nb_P \leftarrow nb_P + 1$
    **else**
        $total_N \leftarrow total_N + (max_N \cdot w_i)$
        $nb_N \leftarrow nb_N + 1$
    **end if**
**end for**
$avg_P(M) \leftarrow total_P / nb_P$
$avg_N(M) \leftarrow total_N / nb_N$
**if** $avg_P(M) \geq avg_N(M)$ **then**
    $P \leftarrow +1$
**else**
    $P \leftarrow -1$
**end if**
**return** $P$

In order to apply the ABSim approach to our running example, we use a simple similarity measure that consists on computing the number of common symbols between the sequences. The first iteration computes the common symbols between the instance $Q_1$ of the query bag and the four related instances $B_{11}$, $B_{21}$, $B_{31}$ and $B_{41}$ and stores the results in the first column of the matrix $M$.

$$\mathbf{M} = \begin{pmatrix} \mathbf{4} & \mathbf{-} & \mathbf{-} \\ \mathbf{4} & \mathbf{-} & \mathbf{-} \\ \mathbf{5} & \mathbf{-} & \mathbf{-} \\ \mathbf{3} & \mathbf{-} & \mathbf{-} \\ \mathbf{-} & \mathbf{-} & \mathbf{-} \end{pmatrix} \begin{matrix} B_1 \\ B_2 \\ B_3 \\ B_4 \\ B_5 \end{matrix}$$

After performing the third iteration, all the columns of the matrix $M$ are computed.

$$\mathbf{M} = \begin{pmatrix} 4 & 5 & 0 \\ 4 & 4 & - \\ 5 & 3 & - \\ 3 & 3 & 3 \\ - & 6 & 2 \end{pmatrix} \begin{matrix} B_1 \\ B_2 \\ B_3 \\ B_4 \\ B_5 \end{matrix}$$

Using the SMS aggregation method, we have the following results:
$total_P (M) = 0$
$total_N (M) = 14$
The query bag Q is finally classified as negative. In order to use the WAMS aggregation method, we need to specify a weight value for each instance. We suppose that all sequences are equally weighted, then we have the following results:
$avg_P (M) = 0$
$avg_N (M) = 4.66$
The query bag *Q* is finally classified as negative.

# 5. Experiments

We applied the naive approach, ABClass and ABSim to solve the problem of IRR prediction in bacteria. Ionizing-radiation-resistant bacteria (IRRB) are important in biotechnology. They could be used for bioremediation of radioactive wastes and in the therapeutic industry. The proposed MIL-based prediction systems aim to affiliate an unknown bacterium to either IRRB or ionizing-radiation-sensitive bacteria (IRSB).

## 5.1. Experimental environment

For our tests, we used the dataset described in (Aridhi *et al*. 2016). This dataset consists of 28 bags. Each bag/bacterium contains 25 to 31 instances that correspond to proteins implicated in basal DNA repair in IRRB. Bacteria represent the bags and primary structures of basal DNA repair proteins represent the instances. The used across bag relation is the orthology. Orthologous genes are assumed to have the same biological functions in different species. In the preprocessing step, we used the perfectBLAST (Santiago-Sotelo and Ramirez-Prado 2012) tool to identify orthologous proteins. We note that some proteins do not have any ortholog in some bags. We do not have the same number of instances in each bag.

Additional and more detailed information about our datasets and our experiments can be found in the following link: *http://homepages.loria.fr/SAridhi/software/MIL/* . We used WEKA (Hall *et al*. 2009) data mining tool in order to apply existing well-known classifiers to test the proposed approaches. The similarity measure used when applying the ABSim approach is the local alignment score. We used BLAST (Altschul et al. 1990) to compute this score.

## 5.2. Experimental protocol

Experiments were carried out using the LOO technique. In order to evaluate ABClass and the naive approach, we need to encode the protein sequences of each bag using a set of features/motifs. We generated these motifs using the DMS motif extraction method (Maddouri and Elloumi, 2004). DMS allows building motifs that can discriminate a family of proteins from other ones. It first identifies motifs in the protein sequences. Then, the extracted motifs are filtered in order to keep only the discriminative and minimal ones. A substring is considered to be discriminative between the family *F* and the other families if it appears in *F* significantly more than in the other families. DMS extracts discriminative motifs according to $\alpha$ and $\beta$ thresholds where $\alpha$ is the minimum rate of motif occurrences in the sequences of a family *F* and $\beta$ is the maximum rate of motif occurrences in all sequences except those of the family *F*. In the following, we present the used motif extraction settings according to the values of $\alpha$ and $\beta$:

- *S1 ($\alpha = 1$ and $\beta = 0.5$)*: used to extract frequent motifs with medium discrimination.
- *S2 ($\alpha = 1$ and $\beta = 1$)*: used to extract frequent motifs without discrimination.

- *S3 (α = 0.5 and β = 1)*: used to extract motifs having medium frequencies without discrimination.
- *S4 (α = 0 and β = 1)*: used to extract infrequent and non discriminative motifs.
- *S5 (α = 1 and β = 0)*: used to extract frequent and strictly discriminative motifs.

## 5.3. Experimental results

Table 1 presents, for each value of *α* and *β*, the number of extracted motifs from each set of related sequences (i.e., orthologous proteins). For the setting *S5*, there are no frequent and strictly discriminative motifs found for most proteins. That is why we will not use these values of *α* and *β* for our next experiments. We note that the number of extracted motifs increases for high values of *β* and low values of *α*. As presented in Table 1, the number of infrequent and non discriminative motifs is very high. When applying the naive approach, we first encode the data using the union of the extracted motifs from each set of related sequences. We show in Table 2 the sparsity of the attribute-value matrix which measures the fraction of zero elements over the total number of elements. The sparsity is generally proportional to the number of used motifs. For example, it goes from 84% with 1141 motifs to 93.5% with 7670 motifs.

***Table 1 Number of extracted motifs for each set of orthologous proteins using a minimum motif length = 3***

| Protein ID | Motif extraction setting | | | | |
|---|---|---|---|---|---|
| | *S1* | *S2* | *S3* | *S4* | *S5* |
| *P1* | *348* | *352* | *612* | *2226* | *229* |
| *P2* | *15* | *76* | *1139* | *5152* | *0* |
| *P3* | *6* | *41* | *681* | *4361* | *0* |
| *P4* | *2* | *21* | *446* | *3751* | *0* |
| *P5* | *1* | *1* | *119* | *1698* | *0* |
| *P6* | *11* | *29* | *349* | *3379* | *0* |
| *P7* | *5* | *18* | *371* | *3907* | *1* |
| *P8* | *3* | *62* | *484* | *3910* | *0* |
| *P9* | *7* | *42* | *780* | *4211* | *0* |
| *P10* | *25* | *90* | *719* | *3830* | *0* |
| *P11* | *3* | *7* | *200* | *2769* | *0* |
| *P12* | *4* | *17* | *144* | *1871* | *0* |
| *P13* | *0* | *1* | *111* | *1544* | *0* |
| *P14* | *2* | *12* | *133* | *2444* | *0* |
| *P15* | *3* | *50* | *303* | *2071* | *0* |
| *P16* | *0* | *1* | *187* | *2659* | *0* |
| *P17* | *3* | *27* | *349* | *2712* | *0* |
| *P18* | *0* | *1* | *81* | *1752* | *0* |
| *P19* | *7* | *14* | *427* | *3800* | *0* |
| *P20* | *2* | *20* | *343* | *3218* | *0* |
| *P21* | *21* | *79* | *882* | *4581* | *1* |
| *P22* | *18* | *173* | *785* | *3910* | *1* |
| *P23* | *5* | *43* | *524* | *4152* | *0* |
| *P24* | *5* | *48* | *520* | *3861* | *0* |
| *P25* | *1* | *5* | *264* | *2563* | *0* |
| *P26* | *22* | *72* | *778* | *3355* | *2* |
| *P27* | *5* | *9* | *162* | *1667* | *0* |
| *P28* | *16* | *111* | *572* | *3308* | *1* |
| *P29* | *2* | *11* | *189* | *2729* | *0* |
| *P30* | *9* | *66* | *281* | *1852* | *0* |
| *P31* | *0* | *0* | *92* | *2061* | *0* |
| Total | *551* | *1499* | *13027* | *95304* | *235* |

*Table 2 Sparsity of the attribute-value matrix generated for the naive approach*

| Motif extraction setting | Total number of motifs | Sparsity (%) |
|---|---|---|
| *S1* | 519 | 84.3 |
| *S2* | 1141 | 84 |
| *S3* | 4167 | 89.6 |
| *S4* | 7670 | 93.5 |

Both ABClass and ABSim approaches provide good overall accuracy results compared to those obtained using the naive approach (see Figure 3). This shows that our proposed approaches are efficient. The results provided by ABSim using the SMS aggregation method are slightly better than those obtained using WAMS. The best result is reached using ABClass approach and the motif extraction settings S1, S2 and S3. Using these three settings, a minimum threshold of frequency and/or discrimination should be reached when extrcating motifs. The figures Figure 3 (a) and Figure 3 (b) show the impact of the motif extraction settings on the prediction results using the naive approach and ABClass. For example, using MISVM classifier, the accuracy varies from 53.5% using S1 to 82.1% using S3. Although the motifs extracted using S1 are discriminative, the naive approach does not provide good accuracy results for most multiple instance classifiers. For some classifiers, the results using S1 are the lowest comparing with the other motif extraction settings. However, using this setting, ABClass provides good results since it reaches 100% of accuracy using SVM, SMO and IBk classifiers, 96.4% using Logistic and 93.3% using J48. This could be explained by the fact that the naive approach looses the advantage of representing the instances using discriminative motifs when it uses the union of all motifs in the data encoding step. Using S4, ABClass does not reach 100% of accuracy although it succeeds to reach it with some classifiers using the other three settings S1, S2 and S3. No constraints related to frequency ($\alpha = 0$) or discrimination ($\beta = 1$) were required when extracting motifs using S4.

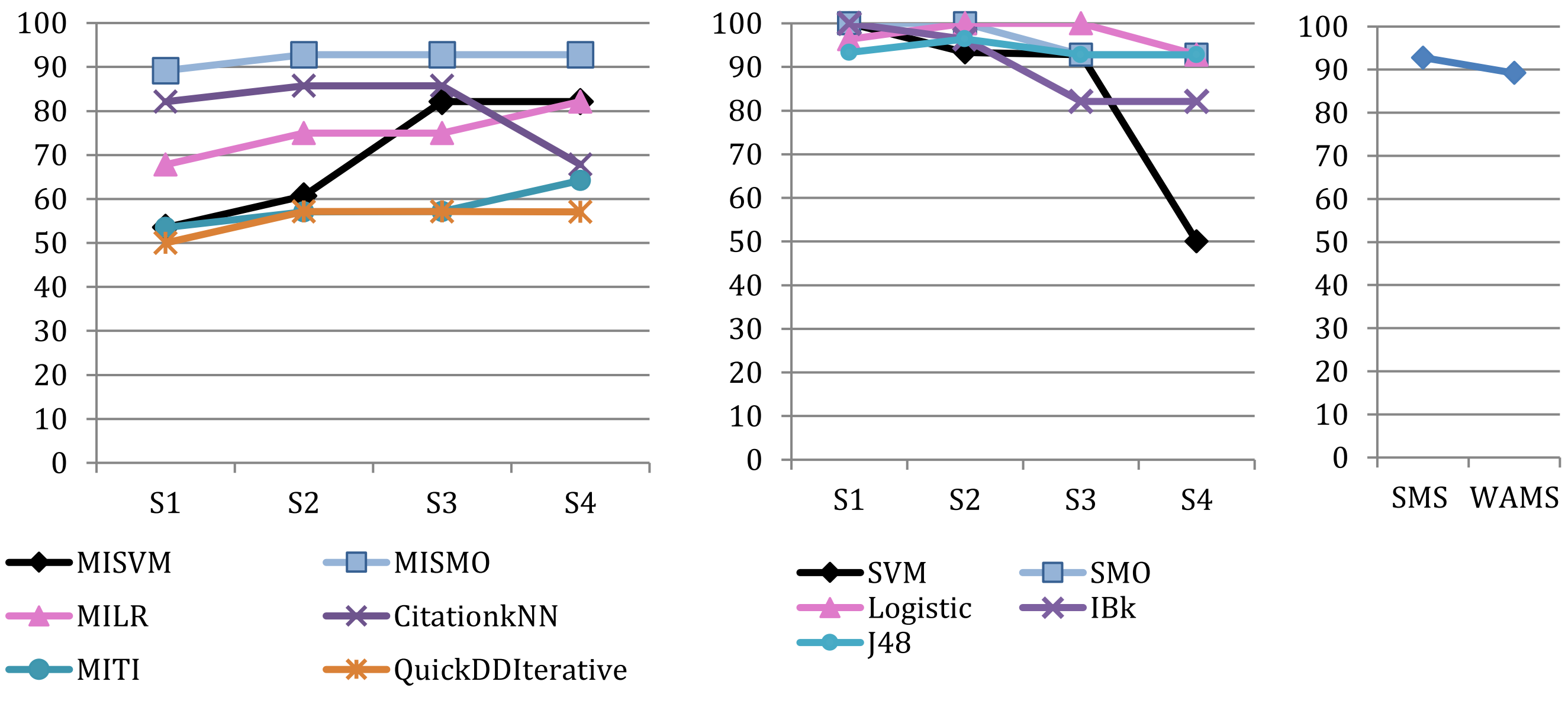

(a) Naive approach (b) ABClass (c) ABSim

***Figure 3 Accuracy results***

We compute the rate of classification models that contribute to predict the true class of each bacterium using ABClass approach (see Table 3). We present this rate for the motif extraction setting that provides the best accuracy rates i.e., *S1* and the setting that provides low accuracy pourcentages, i.e., *S4*. The rate of successful classification models that does not exceed 60% are marked with bold text. The two bacteria *B11* and *B15* often generate low rates. We note that the results are similar to those found in (Aridhi *et al*. 2016). Although *B11* is sometimes successfully classified, its higher successful classification rate does not exceed 62.5%. The rate of B15 does not reach 30% using S4. These results may help to understand some characteristics of the studied bacteria, in particular *M*. radiotolerans (*B11*) and B. abortus (*B15*) that present the lowest rates. A possible biological explanation is provided in (Zoghlami *et al*. 2018; Aridhi *et al*. 2016) and notes that this could be explained by the increased rate of sequence evolution in endosymbiotic bacteria (Woolfit and Bromham 2003).

***Table 3 Rate of successful classification models using ABClass approach and LOO evaluation method***

| Bacterium | S1 motif extraction setting | | | | | S4 motif extraction setting | | | | |
|---|---|---|---|---|---|---|---|---|---|---|
| ID | SVM | SMO | Logistic | IBk | J48 | SVM | SMO | Logistic | IBk | J48 |
| *B1* | 86.3 | 86.3 | 90.9 | 90.9 | 81.8 | **24** | 68 | 80 | **44** | 60 |
| *B2* | 96.2 | 96.2 | 96.2 | 96.2 | 96.2 | 61.2 | 100 | 100 | 100 | 96.7 |
| *B3* | 92.5 | 92.5 | 92.5 | 92.5 | 92.5 | 61.2 | 100 | 100 | 100 | 90.3 |
| *B4* | 96.1 | 96.1 | 96.1 | 96.1 | 92.3 | 66.6 | 100 | 100 | 100 | 93.3 |
| *B5* | 100 | 100 | 100 | 100 | 92.3 | **53.3** | 100 | 100 | 100 | 86.6 |
| *B6* | 100 | 100 | 100 | 100 | 92.3 | **46.6** | 100 | 100 | 100 | 86.6 |
| *B7* | 88.8 | 88.8 | 92.5 | 92.5 | 88.8 | **58** | 100 | 100 | 100 | 93.5 |
| *B8* | 92 | 92 | 92 | 92 | 92 | **41.3** | 100 | 100 | 96.5 | 93.1 |
| *B9* | 95.6 | 92 | 91.3 | 91.3 | 86.9 | **36** | 100 | 100 | 96 | 84 |
| *B10* | 88 | 100 | 88 | 88 | 84 | **32.1** | 100 | 100 | 92.8 | 82.1 |
| ***B11*** | **54.1** | 62.5 | **45.8** | **45.8** | **41.6** | **14.2** | **17.8** | **46.4** | **10.7** | **46.4** |
| *B12* | 91.6 | 91.6 | 91.6 | 91.6 | 91.6 | **42.8** | 100 | 100 | 100 | 92.8 |
| *B13* | 95.6 | 95.6 | 95.6 | 95.6 | 82.6 | **25.9** | 92.5 | 96.2 | **18.5** | 66.6 |
| *B14* | 84 | 80.7 | 84.6 | 84.6 | 61.5 | **33.3** | 96.6 | 96.2 | **43.3** | 70 |
| ***B15*** | 83.3 | 83.3 | 87.5 | 87.5 | 79.1 | **17.8** | **10.7** | **3.5** | **10.7** | **28.5** |
| *B16* | 100 | 100 | 100 | 100 | 100 | 80 | 100 | 100 | 100 | 100 |
| *B17* | 95.8 | 95.8 | 95.8 | 95.8 | 95.8 | 81.4 | 96.2 | 96.2 | 100 | 96.2 |
| *B18* | 100 | 100 | 100 | 100 | 100 | 68 | 100 | 100 | 100 | 100 |
| *B19* | 100 | 100 | 100 | 100 | 100 | 65.3 | 100 | 100 | 100 | 100 |
| *B20* | 88 | 96 | 92 | 92 | 88 | **51.7** | 86.2 | 89.6 | 93.1 | **58.6** |
| *B21* | 100 | 100 | 100 | 100 | 100 | **55.1** | 93.1 | 93.1 | 93.1 | 82.7 |
| *B22* | 96.1 | 96.1 | 96.1 | 96.1 | 96.1 | 80 | 100 | 100 | 100 | 100 |
| *B23* | 88.8 | 92.5 | 96.2 | 96.2 | 92.5 | **48.3** | 100 | 100 | 100 | 96.7 |
| *B24* | 88.8 | 92.5 | 96.2 | 96.2 | 92.5 | **48.3** | 100 | 100 | 100 | 100 |
| *B25* | 88.8 | 92.5 | 96.2 | 96.2 | 92.5 | **48.3** | 100 | 100 | 100 | 96.7 |
| *B26* | 88.8 | 92.5 | 96.2 | 96.2 | 92.5 | **48.3** | 100 | 100 | 100 | 100 |
| *B27* | 100 | 100 | 100 | 100 | 100 | 62.9 | 100 | 100 | 100 | 96.2 |
| *B28* | 96.1 | 96.1 | 96.1 | 96.1 | 96.2 | 66.6 | 96.6 | 100 | 96.6 | 96.6 |

# 6. Conclusion

In this paper, we addressed the issue of MIL in the case of sequence data. We focused on data that present relationships between instances of different bags. We described two novel approaches for MIL in sequence data with across bag dependencies. In the first method, called ABClass, each set of related instances will be presented by its own motifs vector. A discriminative classifier is then applied to compute a partial classification result. The second method, called ABSim, uses a matrix to store similarity measurement score vectors to discriminate the related instances. Both methods apply an aggregation step in order to generate the final classification result. We applied ABClass and ABSim to solve the problem of IRR prediction in bacteria. By running experiments, we have shown that the proposed approaches are efficient. A better accuracy result could be provided by ABClass according to the used settings. In the future work, we will study how the use of a priori knowledge can improve the efficiency of the presented approaches. We specifically want to define weights for sequences by using a priori knowledge in the learning phase. Another avenue is to tackle this problem through protein 3D structure based on the work reported in (Dhifli *et al*. 2014).